\pdfoutput=1

\documentclass[11pt]{article}

\usepackage[]{ACL2023}

\usepackage{times}
\usepackage{latexsym}

\usepackage[T1]{fontenc}

\usepackage[utf8]{inputenc}

\usepackage{microtype}

\usepackage{inconsolata}

\usepackage{bm}
\usepackage{amsmath}
\usepackage{url}
\usepackage{graphicx}
\usepackage{float} 
\usepackage{subfigure}
\usepackage{multirow}
\usepackage{makecell}
\usepackage{xpinyin}
\usepackage{enumerate}

\usepackage{xcolor}
\usepackage{framed}
\usepackage{color}
\usepackage{bbding}
\usepackage{soul}
\usepackage{colortbl}
\newcolumntype{L}[1]{>{\raggedright\let\newline\\\arraybackslash\hspace{0pt}}m{#1}}
\setul{2pt}{2pt}
\usepackage{booktabs}
\usepackage{amsmath,amsfonts}
\usepackage[utf8]{inputenc}
\usepackage{cleveref}

\usepackage{ulem}
\usepackage{diagbox}
\crefname{section}{§}{§§}
\Crefname{section}{§}{§§}
\usepackage{enumitem}
\setenumerate[1]{itemsep=0pt,partopsep=0pt,parsep=\parskip,topsep=5pt}
\setitemize[1]{itemsep=0pt,partopsep=0pt,parsep=\parskip,topsep=5pt}
\setdescription{itemsep=0pt,partopsep=0pt,parsep=\parskip,topsep=5pt}
\usepackage{arydshln}

\usepackage[linesnumbered,ruled,vlined]{algorithm2e}

\title{MM-Sum: Large-Scale Multilingual Multimodal Abstractive Summarization for 44 Languages?}
\title{Towards Making the Most of Vision \\ for Multimodal Abstractive Summarization on 44 Languages}
\title{Towards Making the Most of Vision \\ for Multimodal Abstractive Summarization}
\title{Summary-Oriented Vision Modeling for Multimodal Abstractive Summarization}

\author{
  Yunlong Liang\textsuperscript{1}\thanks{ \ \ Work was done when Liang and Wang was interning at Pattern Recognition Center, WeChat AI, Tencent Inc, China.}  , 
  Fandong Meng\textsuperscript{2}, 
  \textbf{Jinan Xu}\textsuperscript{1}\thanks{ \ \ Jinan Xu is the corresponding author.}  , 
  Jiaan Wang\textsuperscript{2}, 
  \textbf{Yufeng Chen}\textsuperscript{1}
   and \textbf{Jie Zhou}\textsuperscript{2}\\
  \textsuperscript{1}Beijing Key Lab of Traffic Data Analysis and Mining, \\Beijing Jiaotong University, Beijing, China \\
  \textsuperscript{2}Pattern Recognition Center, WeChat AI, Tencent Inc, China \\
  \texttt{\{yunlongliang,jaxu\}@bjtu.edu.cn} \\
  \texttt{fandongmeng@tencent.com} \\
}

\begin{document}
\maketitle
\begin{abstract}
Multimodal abstractive summarization (MAS) aims to produce a concise summary given the multimodal data (text and vision). Existing studies mainly focus on how to effectively use the visual features from the perspective of an article, having achieved impressive success on the high-resource English dataset. However, less attention has been paid to the visual features from the perspective of the summary, which may limit the model performance, especially in the low- and zero-resource scenarios. In this paper, we propose to improve the summary quality through summary-oriented visual features. To this end, we devise two auxiliary tasks including \emph{vision to summary task} and \emph{masked image modeling task}. Together with the main summarization task, we optimize the MAS model via the training objectives of all these tasks. By these means, the MAS model can be enhanced by capturing the summary-oriented visual features, thereby yielding more accurate summaries. 
Experiments on 44 languages, covering mid-high-, low-, and zero-resource scenarios, verify the effectiveness and superiority of the proposed approach, which achieves state-of-the-art performance under all scenarios. Additionally, we will contribute a large-scale multilingual multimodal abstractive summarization (MM-Sum) dataset.\footnote{The code and data are publicly available at: \url{https://github.com/XL2248/SOV-MAS}.} 

\end{abstract}

\section{Introduction}
\label{intro}
Given an article and several images as inputs, as shown in~\autoref{fig.1}, multimodal abstractive summarization (MAS)~\cite{sanabria18how2,li-etal-2017-multi,li2018multi,zhu-etal-2018-msmo,jangra2020multi} aims to generate a concise textual summary, which can help people quickly grasp the core information. Therefore, MAS has widespread application and attracts increasing attention with the rapid proliferation of multimedia content~\cite{apostolidis2021video,feng-etal-2022-msamsum,qiu2022mhms}. 
\textbf{\begin{figure}[t]
    \centering
    \includegraphics[width=0.46\textwidth]{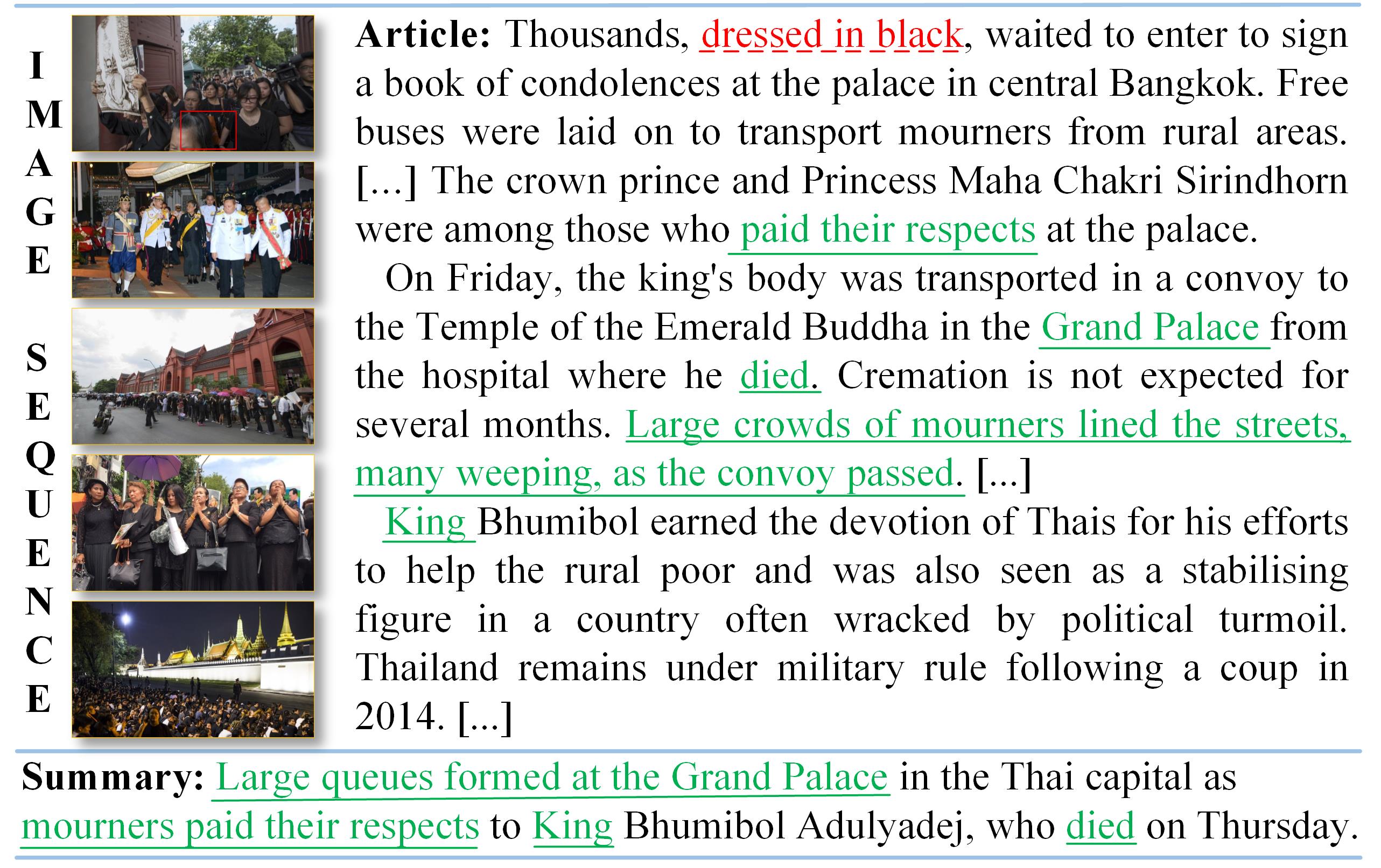}
    \caption{An example of our MM-Sum dataset. Inputs: an article and image sequence pair; Output: summary. As we can see, the image sequence also concisely paraphrases the summary. The \textcolor[RGB]{255,59,59}{\dashuline{red}} content indicates its associated object is useless to the summary while the \textcolor[RGB]{0,176,80}{\uline{green}} counterparts represent important information. } 
    \label{fig.1}\vspace{-15pt}
\end{figure}}

Recently, many studies have been carried out to effectively inject the visual features into MAS models~\cite{li2018read,li-etal-2020-vmsmo,zhu2020multimodal,zhu2021graph,zhang2021unims,zhang2021hierarchical,palaskar-etal-2019-multimodal,liu-etal-2020-multistage,yu-etal-2021-vision}. For instance,~\citet{palaskar-etal-2019-multimodal} and~\citet{zhang2021hierarchical} explore the hierarchy between the textual article and visual features, and integrate them into the MAS model.~\citet{liu-etal-2020-multistage} design a multistage fusion network to model the fine-grained interactions between the two modalities. And~\citet{yu-etal-2021-vision} study multiple multimodal fusion methods to infuse the visual features into generative pre-trained language models, \emph{e.g.}, BART~\cite{lewis-etal-2020-bart}. Despite their success on the high-resource English dataset, they only model visual features from the perspective of an article and neglect the relevance of visual features to the summary, which restricts their potential performance especially on the training dataset with limited scale. For example, though the object ``black clothes'' in the first image of~\autoref{fig.1} is associated with the article content (\textcolor[RGB]{255,59,59}{\dashuline{red}} part), the object contributes little to the summary. Thus, the MAS model should focus on summary-oriented visual features. However, the visual features are generally implicitly learned via the MAS objective, which cannot help the model learn to explicitly discard the needless visual information.   

To address this issue, in this paper, we propose a \textbf{S}ummary-\textbf{O}riented \textbf{V}ision enhanced MAS (SOV-MAS) training framework to generate more accurate summaries through explicitly improving the relevance of visual features to the summary. To this end, we design two summary-oriented vision modeling tasks, namely \emph{vision to summary task}, and \emph{masked image modeling task}. Specifically, as shown in~\autoref{fig.2}, (1) the \emph{vision to summary task} is to produce the concise summary by only taking the image sequence; (2) the \emph{masked image modeling task} aims to predict the semantic class distribution of the regions in one fully masked image given the summary and the remaining images. Together with the main multimodal summarization task, the MAS model is optimized through the joint objectives of all these tasks. In this way, the model is enhanced to explicitly exploit the summary-oriented visual features, thus leading to more accurate summaries. 

To validate the SOV-MAS framework on various languages and diverse settings, we construct the first large-scale \textbf{M}ultilingual \textbf{M}ultimodal \textbf{Sum}marization dataset (MM-Sum) based on XL-Sum~\cite{hasan-etal-2021-xl}, a multilingual summarization dataset. The MM-Sum covers 44 languages with mid-high-, low- and zero-resource scenarios. Experiments on these settings show that our model significantly outperforms related methods in terms of ROUGE~\cite{lin-2004-rouge} scores, especially under the low- and zero-resource settings, demonstrating its effectiveness. Besides, we extend our approach to two previous best MAS models (\emph{i.e.}, VG-BART and VG-T5~\cite{yu-etal-2021-vision}). Human evaluation and the results on How2~\cite{sanabria18how2} benchmark further suggest the superiority and generalizability of our approach. In summary, our main contributions are:

\begin{itemize}[leftmargin=*]

\item To the best of our knowledge, we are the first that contributes a large-scale multilingual multimodal summarization dataset (44 languages, 1.1M article-summary pairs with 3.5M images). 

\item We propose two general summary-oriented vision modeling tasks, which substantially boost the summary quality and are flexible and easy to be extended to existing MAS models. 

\item Experiments on MM-Sum show that our model builds new state-of-the-art performance in all scenarios, especially on the low and zero resource where the fewer the data are (mid-high$\rightarrow$low$\rightarrow$zero), the greater the improvement we gain. Besides, results on the How2 dataset show the generalizability of our approach. 
\item When jointly training the MAS model on multiple languages, we find that our model learns transferable visual features among languages, where the vision serves as an anchor in the zero-resource languages.
\end{itemize}

\section{Background}\label{bg}


\subsection{Problem Formulation}
\label{pf}
Given an input article $\mathcal{X}$$=$$\{x_k\}_{k=1}^{|\mathcal{X}|}$ and the corresponding object sequence $\mathcal{O}$$=$$\{o_{ij}\}_{i=1,j=1}^{i\leq n,j\leq m}$, where $x_k$ denotes the $k$-th token and $o_{ij}$ represents the detected $j$-th object of the $i$-th image ($n$, $m$ is the number of images and detected objects in each image, respectively), the MAS task is defined as:
\begin{equation}\nonumber
\setlength{\abovedisplayskip}{5pt}
\setlength{\belowdisplayskip}{5pt}
\label{eq:ms}
    p(\mathcal{Y}|\mathcal{X}, \mathcal{O}) = \prod_{t=1}^{|\mathcal{Y}|}p(y_t|\mathcal{X}, \mathcal{O}, y_{<t}),
\end{equation}
where $y_{<t}$ indicates the tokens before the $t$-th time step in the summary $\mathcal{Y}$$=$$\{y_t\}_{t=1}^{|\mathcal{Y}|}$.
\textbf{\begin{figure*}[t]
    \centering
    \includegraphics[width=0.94\textwidth]{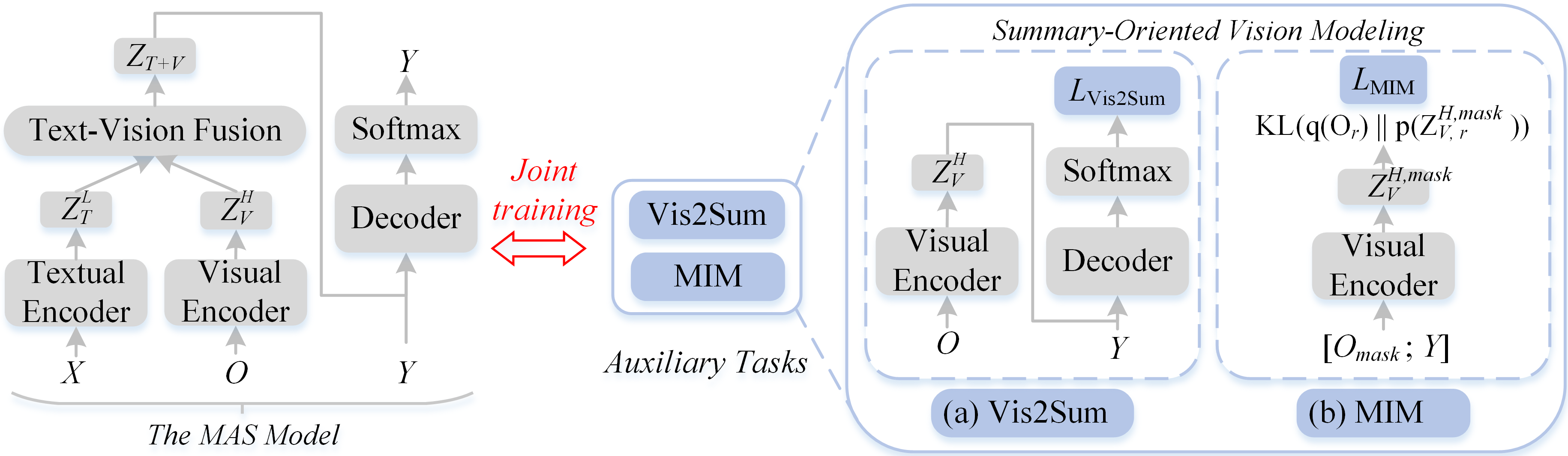}
    \caption{The overview of our model architecture. The left part is a general MAS model, which is enhanced by two summary-oriented vision modeling tasks. As shown in the right part, the two auxiliary tasks including (a) \emph{vision to summary task} (Vis2Sum) and (b) \emph{masked image modeling task} (MIM), are proposed to focus on the summary-oriented visual features and thus benefit the multimodal summarization task. }
    \label{fig.2}\vspace{-12pt}
\end{figure*}}
\subsection{The MAS Model}
\label{MAS}
Based on the pre-trained language models (\emph{e.g.}, BART), \citet{yu-etal-2021-vision} design a variant of transformer~\cite{vaswani2017attention} with four modules: textual encoder, visual encoder, text-vision fusion, and decoder, as shown in the left part of~\autoref{fig.2}, which achieves good performance on MAS. 

\noindent \textbf{Textual Encoder.} The input text $\mathcal{X}$ is firstly tokenized and mapped to a sequence of token embeddings $\mathbf{X}$. Then, the positional encodings $\mathbf{E}_{pe}$ are pointwisely added to $\mathbf{X}$ to keep the positional information~\citep{vaswani2017attention}:
\begin{equation}\label{input_embed}\nonumber
\setlength{\abovedisplayskip}{5pt}
\setlength{\belowdisplayskip}{5pt}
\resizebox{0.89\hsize}{!}{$
\begin{split}
\mathbf{Z}_{T}^{0} = \mathbf{X} + \mathbf{E}_{pe}, \ \{\mathbf{Z}_{T}^{0}, \mathbf{X}, \mathbf{E}_{pe}\} \in \mathbb{R}^{|\mathcal{X}| \times d},
\end{split}
$}
\end{equation}
where $d$ is the feature dimension. It forms the input features $\mathbf{Z}_{T}^{0}$ to the encoder, which consists of $L$ stacked layers and each layer includes two sub-layers: 1) Multi-Head Attention ($\mathrm{MHA}$) and 2) a position-wise Feed-Forward Network ($\mathrm{FFN}$):
\begin{equation}
\setlength{\abovedisplayskip}{5pt}
\setlength{\belowdisplayskip}{5pt}
\resizebox{0.89\hsize}{!}{$
\begin{split}
    \mathbf{S}^\ell_{T} &= \mathrm{MHA}(\mathbf{Z}^{\ell-1}_{T}) + \mathbf{Z}^{\ell-1}_{T},\ \mathbf{S}^\ell_{T} \in \mathbb{R}^{|\mathcal{X}| \times d},\nonumber\\
    \mathbf{Z}^\ell_{T} &= \mathrm{FFN}(\mathbf{S}^\ell_{T}) + \mathbf{S}^\ell_{T},\ \mathbf{Z}^\ell_{T} \in \mathbb{R}^{|\mathcal{X}| \times d},\nonumber
\end{split}
$}
\end{equation}
where $\mathbf{Z}^{\ell}_{T}$ is the state of the $\ell$-th encoder layer.

\noindent \textbf{Visual Encoder.} Following~\citet{yu-etal-2021-vision,zhang2021hierarchical,zhang2021unims,liang2020infusing,liang-etal-2022-msctd,LIANG2022103714}, the object sequence $\mathcal{O}$ is extracted from the image by the Faster R-CNNs~\cite{NIPS2015_14bfa6bb} (actually, we have several images instead of only one image, please refer to~\autoref{sec:vision_features} for details). Then the visual features are fed into the visual encoder with $H$ layers. Finally, we obtain the output visual features $\mathbf{Z}_V^{H}$:
\begin{equation}
\setlength{\abovedisplayskip}{5pt}
\setlength{\belowdisplayskip}{5pt}
\resizebox{0.89\hsize}{!}{$
\begin{split}
    \mathbf{S}^h_{V} &= \mathrm{MHA}(\mathbf{Z}^{h-1}_{V}) + \mathbf{Z}^{h-1}_{V},\ \mathbf{S}^h_{V} \in \mathbb{R}^{|\mathcal{O}| \times d_v},\nonumber\\
    \mathbf{Z}^h_{V} &= \mathrm{FFN}(\mathbf{S}^h_{V}) + \mathbf{S}^h_{V},\ \mathbf{Z}^h_{V} \in \mathbb{R}^{|\mathcal{O}| \times d_v},\nonumber
\end{split}
$}
\end{equation}
where $\mathbf{Z}^h_{V}$ is the extracted visual features $\mathbf{O}$.

\noindent \textbf{Text-Vision Fusion.} The fusion method is vision-guided multi-head attention. Firstly, the query $\mathbf{Q}$ is linearly projected from the textual features $\mathbf{Z}_{T}^L$, and the key $\mathbf{K}$ and value $\mathbf{V}$ are linearly projected from the visual features $\mathbf{Z}_V^H$. Secondly, a Cross-modal Multi-Head Attention ($\mathrm{CMHA}$) is applied to get the text queried visual features $\mathbf{M}$. Then, a forget gate $\mathbf{G}$ is used to filter redundant and noisy information from the visual features. Finally, we obtain the vision-guided output $\mathbf{Z}_{T+V}$ by concatenating the textual features $\mathbf{Z}_{T}^L$ and the result of a point-wise multiplication $\mathbf{G} \otimes \mathbf{M}$, and then linearly project it to the original dimension $d$. Formally, the text-vision fusion process is:
\begin{equation}\nonumber
\setlength{\abovedisplayskip}{5pt}
\setlength{\belowdisplayskip}{5pt}
\resizebox{0.89\hsize}{!}{$
\begin{split}
    \mathbf{Q} &= \mathbf{Z}_{T}^L \mathbf{W}_q, \ \mathbf{Q} \in \mathbb{R}^{|\mathcal{X}| \times d_c},\\
    \mathbf{K} &= \mathbf{Z}_V^H \mathbf{W}_k, \ \mathbf{V} = \mathbf{Z}_V^H \mathbf{W}_v, \  \mathbf{K}, \mathbf{V} \in \mathbb{R}^{|\mathcal{O}| \times d_c},\\
    \mathbf{M} &= \mathrm{CMHA}(\mathbf{Q},\mathbf{K},\mathbf{V}), \ \mathbf{M} \in \mathbb{R}^{|\mathcal{X}| \times d_c}, \\
    \mathbf{G} &= \mathrm{Sigmoid}(\mathrm{Concat}(\mathbf{Z}_{T}^L, \mathbf{M}) \mathbf{W}_g + \mathbf{b}_g), \\
    \mathbf{Z}_{T+V} &= \mathrm{Concat}(\mathbf{Z}_{T}^L, \mathbf{G} \otimes \mathbf{M}) \mathbf{W}_z + \mathbf{b}_z,
\end{split}
$}
\end{equation}
where $\mathrm{Concat}$ is the concatenation operation and $\mathbf{W}_*$ and $\mathbf{b}_*$ are trainable weights.

\noindent \textbf{Decoder.} Similar to the encoder, but each of $L$ decoder layers includes an additional Multi-Head Cross-Attention sub-layer ($\mathrm{MHCA}$):
\begin{equation}
\setlength{\abovedisplayskip}{5pt}
\setlength{\belowdisplayskip}{5pt}
\resizebox{0.89\hsize}{!}{$
\begin{split}
\label{eq:decoder}
    \mathbf{S}^\ell_{dec} &= \mathrm{MHA}(\mathbf{Z}^{\ell-1}_{dec}) + \mathbf{Z}^{\ell-1}_{dec},\ \mathbf{S}^{\ell-1}_{dec}\in \mathbb{R}^{|\mathcal{Y}| \times d},\\
    \mathbf{C}^\ell_{dec} &= \mathrm{MHCA}(\mathbf{S}^{\ell}_{dec}, \mathbf{Z}_{T+V}) + \mathbf{S}^\ell_{dec}, \\
    \mathbf{Z}^\ell_{dec} &= \mathrm{FFN}(\mathbf{C}^\ell_{dec}) + \mathbf{C}^\ell_{dec},\ \mathbf{C}^\ell_{dec} \in \mathbb{R}^{|\mathcal{Y}| \times d},
\end{split}
$}
\end{equation}
where $\mathbf{Z}^{\ell}_{dec} \in \mathbb{R}^{|\mathcal{Y}| \times d}$ denotes the state of the $\ell$-th decoder layer. Then, at each decoding time step $t$, the top-layer ($L$-th) decoder hidden state $\mathbf{Z}^{L}_{dec,t}$ is fed into the softmax layer to produce the probability distribution of the next target token as:
\begin{equation}
\setlength{\abovedisplayskip}{5pt}
\setlength{\belowdisplayskip}{5pt}
\resizebox{0.89\hsize}{!}{$
\begin{split}
    p(y_{t}|\mathcal{X}, \mathcal{O}, y_{<t}) &= \mathrm{Softmax}(\mathbf{W}_o\mathbf{Z}^{L}_{dec,t}+\mathbf{b}_o),\nonumber
\end{split}
$}
\end{equation}
where $\mathbf{W}_o$ and $\mathbf{b}_o$ are trainable weights. 

Finally, the loss function is formalized as:
\begin{equation}
\setlength{\abovedisplayskip}{5pt}
\setlength{\belowdisplayskip}{5pt}
\resizebox{0.69\hsize}{!}{$
\begin{split} 
\label{eq:cnmt}
    \mathcal{L}_{\text{MAS}} = -\sum_{t=1}^{|\mathcal{Y}|}\mathrm{log}(p(y_{t}|\mathcal{X}, \mathcal{O}, y_{<t})).
\end{split}
$}
\end{equation}

\section{SOV-MAS Framework}
\label{sec:sov-mas}
Based on the vision-guided pre-trained language model described in~\autoref{MAS}, we introduce the proposed \textbf{S}ummary-\textbf{O}riented \textbf{V}ision enhanced MAS ((SOV-MAS)) framework. Specifically, we firstly describe the process of \emph{visual features extraction} in~\autoref{sec:vision_features}. Then, to make the best use of visual features, we design two \emph{summary-oriented vision modeling} tasks in~\autoref{atask}, namely \emph{vision to summary task} and \emph{masked image modeling task}. Finally, we describe the \emph{training and inference} in~\autoref{training_infer}.

\subsection{Visual Features Extraction}
\label{sec:vision_features}
As described in~\autoref{MAS}, there is an image sequence to be extracted by the Faster R-CNNs~\cite{NIPS2015_14bfa6bb} pre-trained on Visual Genome~\cite{krishnavisualgenome}. Specifically, for the $i$-th input image, we obtain a set of detected objects from Faster R-CNNs, \emph{i.e.}, $\mathbf{I}_i$ = $\{\mathbf{v}_{i,1}, \mathbf{v}_{i,2}, \mathbf{v}_{i,3}, ..., \mathbf{v}_{i,m}\}$, where $m$ is the number of extracted objects and $\mathbf{v}_{i,*} \in \mathbb{R}^{d_v}$. Each object is captured by a dense feature representation, which can be mapped back to a bounding box / region (\emph{i.e.}, Region-of-Interest (RoI)). Finally, the image sequence is converted to visual features $\mathbf{I}$$=$$\{\mathbf{v}_{ij}\}_{i=1,j=1}^{i\leq n,j\leq m}$. 

Besides these features from Faster R-CNN, given the fact that Transformer~\cite{vasava-etal-2022-transformer} is becoming popular in computer vision, we experiment with the visual features extracted by the pre-trained Transformer models (\emph{i.e.}, ViT~\cite{dosovitskiy2020image}).

To keep the order information of the image sequence, each image region is encoded as a sum of four types of features~\cite{pmlr-v139-cho21a}:
\begin{equation}\label{input_embed2}\nonumber
\setlength{\abovedisplayskip}{5pt}
\setlength{\belowdisplayskip}{5pt}
\mathbf{o}_{ij} = \mathbf{v}_{ij} + \mathbf{E}^{box}_{ij} + \mathbf{E}^{img}_{i} + \mathbf{E}^{reg}_{j}; i \leq n,j\leq m,
\end{equation}
where $\mathbf{E}^{box}_{ij} \in \mathbb{R}^{d_v}$ denotes RoI bounding box coordinates, which are encoded with a linear layer; $\mathbf{E}^{img}_{i} \in \mathbb{R}^{d_v}$ denotes image id embedding, which is used to discriminate regions from different images; and $\mathbf{E}^{reg}_{j} \in \mathbb{R}^{d_v}$ denotes region id embedding. The image ids and region ids are encoded with learned embeddings~\cite{bert}. The final visual embeddings are denoted as $\mathbf{O}$$=$$\{\mathbf{o}_{ij}\}_{i=1,j=1}^{i\leq n,j\leq m}$. Then, they are fed into the visual encoder for better modeling the intramodal dynamics and enhancing the vision-specific order information.

\subsection{Summary-Oriented Vision Modeling}\label{atask}
We elaborately design two summary-oriented vision modeling tasks, namely \emph{vision to summary task} and \emph{masked image modeling task}, to focus on the summary-oriented visual features. 

\noindent \textbf{Vision to Summary Task (Vis2Sum).}
As illustrated in the right part of \autoref{fig.2} (a), given the object sequence $\mathcal{O}$ extracted from the image sequence, the Vis2Sum task forces the MAS model to directly generate the corresponding summary $\mathcal{Y}$ without seeing the article $\mathcal{X}$. In this manner, the MAS model could acquire the ability to roughly understand the summary and grasp the overall situation. Particularly, we firstly use the visual encoder to encode $\mathcal{O}$, and then use the MAS decoder to predict $\mathcal{Y}$. The training objective of this task can be formulated as:
\begin{equation}
\setlength{\abovedisplayskip}{5pt}
\setlength{\belowdisplayskip}{5pt}
\resizebox{0.85\hsize}{!}{$
\begin{split}
        \mathcal{L}_{\text{Vis2Sum}} &= -\sum_{t=1}^{|\mathcal{Y}|}\mathrm{log}(p(y_{t}|\mathcal{O}, y_{<t})),\\
    p(y_{t}|\mathcal{O}, y_{<t}) &= \mathrm{Softmax}(\mathbf{W}_o\mathbf{Z}^{L,V}_{dec,t}+\mathbf{b}_o),
\end{split}
$}
\end{equation}
where $\mathbf{Z}^{L,V}_{dec,t}$ is the top-layer decoder hidden state at the $t$-th decoding step, while the input of $\mathrm{MHCA}$ is the visual features $\mathbf{Z}_V^H$ instead of $\mathbf{Z}_{T+V}$ in~\autoref{eq:decoder}. 

\noindent \textbf{Masked Image Modeling Task (MIM).}
Our MIM task aims to predict the semantic class distribution of the regions in one fully masked image.  As illustrated in the right part of \autoref{fig.2} (b), for the input of the visual encoder, we firstly mask all regions in one random image (\emph{i.e.}, $m$ objects/regions), which are replaced with zero vectors. Then, we concatenate the masked object sequence $\mathcal{O}_{mask}$ and the summary $\mathcal{Y}$. After feeding the concatenated input [$\mathcal{O}_{mask}$; $\mathcal{Y}$] to the encoder, an MLP classifier is stacked over the output of each masked region to predict the semantic class distribution. Specifically, we denote the predicted class distribution of the $r$-th masked region as p($\mathbf{Z}_{V,r}^{H,mask}$), and use q($\mathbf{O}_r$) to represent the class distribution detected by the Faster R-CNNs~\cite{NIPS2015_14bfa6bb}. The loss function for the MIM is to minimize the KL divergence~\cite{kingma2013auto} of the two class distributions:
\begin{equation}
\setlength{\abovedisplayskip}{5pt}
\setlength{\belowdisplayskip}{5pt}
\resizebox{0.79\hsize}{!}{$
\begin{split}
        \mathcal{L}_{\text{MIM}} &= \sum_{r=1}^{m}\mathrm{D_{KL}}(q(\mathbf{O}_r)||p(\mathbf{Z}_{V,r}^{H,mask})).\\
\end{split}
$}
\end{equation}
Besides, as a variant, we randomly mask regions in the image sequence with a probability of 15\% following previous work~\cite{xing-etal-2021-km}. We denote it as masked region modeling (MRM) and show its effect in~\autoref{tab:abl}.

\subsection{Training and Inference}\label{training_infer}
\noindent \textbf{Monolingual Training.} For monolingual summarization, with the main MAS task and the two auxiliary tasks, the training objective on one specific language is finally formulated as:
\begin{equation}
\setlength{\abovedisplayskip}{5pt}
\setlength{\belowdisplayskip}{5pt}
\resizebox{0.89\hsize}{!}{$
\begin{split}
    &\mathcal{J}_\text{Mono} = \mathcal{L}_{\text{MAS}} + \alpha\mathcal{L}_{\text{Vis2Sum}} + \beta\mathcal{L}_{\text{MIM}},
\end{split}\label{loss_all}
$}
\end{equation}
where $\alpha$ and $\beta$ are balancing factors for the trade-off between $\mathcal{L}_{\text{MAS}}$ and the auxiliary objectives. 

\noindent \textbf{Multilingual Training.} For multilingual summarization, the model can deal with inputs in multiple languages and predict the summary in the corresponding language. Specifically, for each language $l_k$ in the set of $K$ languages $Lang$ = $\{l_1,l_2,...,l_K\}$, the training objective is:
\begin{equation}
\setlength{\abovedisplayskip}{5pt}
\setlength{\belowdisplayskip}{5pt}
\resizebox{0.49\hsize}{!}{$
\begin{split}
    &\mathcal{J}_\text{Multi} = \sum_{k=1}^{K}(\mathcal{J}_\text{Mono}^{l_k}).
\end{split}\label{loss_all_multi}
$}
\end{equation}

During inference, the two auxiliary tasks are not involved and only the MAS model is used to conduct summarization.

\begin{table*}[h]
\begin{minipage}{\textwidth}
\centering
\scalebox{0.72}{
\setlength{\tabcolsep}{1.8mm}{
\begin{tabular}{l|c|c|c||c|c|c}
\hline
&\multicolumn{3}{c||}{\textbf{Monolingual Training}} &\multicolumn{3}{c}{\textbf{Multilingual Training}}  \\ 
\cline{2-4} \cline{5-7}
\textbf{Languages} &\textbf{mT5}&\textbf{VG-mT5} &\textbf{SOV-MAS} (ours) &\textbf{mT5}&\textbf{VG-mT5} &\textbf{SOV-MAS} (ours)\\
\hline
Arabic &33.67/14.06/27.83 &33.88/14.20/28.00 &33.63/13.83/27.64 &34.34/14.30/28.43&33.42/13.58/27.62 &\cellcolor{blue!12}{\uline{34.74/14.48/28.84}} \\
Chinese &40.20/25.39/33.49&39.99/25.19/33.19 &\cellcolor{gray!20}{\uwave{40.59/25.32/33.36}} &40.30/24.97/33.04&40.14/25.29/33.31 &\cellcolor{blue!12}{\uline{41.59/26.52/34.53}}  \\
English &36.99/15.18/29.64 &37.17/14.88/29.41 &37.26/15.02/29.61 &36.65/13.91/28.53&36.62/14.13/28.76 &\cellcolor{blue!12}{\uline{37.86/15.23/29.89}} \\
Hindi &33.66/13.14/27.71&34.82/13.94/28.59 &34.83/13.60/28.25 &35.50/13.91/28.52&35.36/14.16/28.87 &\cellcolor{blue!12}{\uline{36.42/14.95/29.77}}  \\
Indonesian &35.10/15.44/28.91&35.47/15.47/29.12 &35.17/15.35/28.85 &35.84/15.66/29.40&36.50/16.31/30.13 &\cellcolor{blue!12}{\uline{37.50/17.33/31.22}}  \\
Persian &36.14/15.55/29.25&36.12/15.59/29.15 &36.44/15.92/29.50 &36.39/15.84/29.45&36.71/16.19/29.80 &\cellcolor{blue!12}{\uline{37.69/16.90/30.71}} \\
Portuguese &30.13/10.32/22.06&29.69/\ \;9.82/22.10 &29.83/10.05/21.78 &30.84/10.92/22.64&31.22/11.43/23.24   &\cellcolor{blue!12}{\uline{32.32/11.90/23.83}}  \\
Russian &30.01/12.47/24.28&31.38/13.02/25.22 &\cellcolor{gray!20}{\uwave{31.86/13.38/25.45}} &31.12/12.33/24.67&30.42/12.29/24.38  &\cellcolor{blue!12}{\uline{31.96/13.30/25.69}} \\
Spanish &29.51/10.48/22.51&29.50/10.62/22.47 &29.27/10.40/22.43 &29.91/10.70/22.66&30.57/10.96/23.21 &\cellcolor{blue!12}{\uline{31.20/11.64/23.73}}  \\
Tamil &22.31/10.08/20.36&22.30/10.15/20.39  &\cellcolor{gray!20}  {\uwave{22.82/10.55/20.67}} &22.96/10.05/20.75&23.04/10.25/20.94 &\cellcolor{blue!12}{\uline{24.22/10.79/21.92}}  \\
Turkish &30.37/14.39/26.79&30.51/14.41/26.76 &\cellcolor{gray!20}  {\uwave{31.02/14.64/27.20}} &31.93/14.69/27.76&31.44/14.73/27.71 &\cellcolor{blue!12}{\uline{32.94/15.77/29.01}}  \\
Ukrainian &21.57/\ \;8.66/18.64&21.71/\ \;8.89/18.79 &21.84/\ \;8.62/18.69 &22.79/\ \;9.13/19.46&22.60/\ \;9.27/19.55 &\cellcolor{blue!12}{\uline{23.91/\ \;9.97/20.53}}  \\
Urdu &38.22/17.25/31.37&38.07/17.31/31.54 &38.10/16.98/31.18 &38.15/17.12/31.36&38.04/17.32/31.67 &\cellcolor{blue!12}{\uline{39.38/18.38/32.76}} \\
Vietnamese &32.18/15.84/24.83&32.18/15.98/24.84 &32.22/15.99/24.95 &33.71/16.72/25.97&33.78/17.06/26.32 &\cellcolor{blue!12}{\uline{34.78/17.85/27.17}} \\
\cline{1-7}
\textbf{Avg.} &32.14/14.16/26.26  &32.34/14.24/26.39 &32.49/14.26/26.40 &32.88/14.30/26.61 &32.84/14.49/26.82 &\textbf{34.04}/\textbf{15.36}/\textbf{27.83}  \\
\hline
\end{tabular}}}
\caption{The R-1/R-2/R-L results on the mid-high-resource scenario. ``\colorbox{gray!20}{\uwave{*/*/*}}'' and ``\colorbox{blue!12}{\uline{*/*/*}}'' denote statistically significant better than the ``VG-mT5'' model with t-test \emph{p} \textless \ 0.05 and \emph{p} \textless \ 0.01 hereinafter, respectively. The ``Avg.'' indicates average score for each group and the best average scores are \textbf{bold}.}\label{tab:Mid-High}
\end{minipage}
\end{table*}

\section{Experiments}

\subsection{MM-Sum Dataset} There is no multilingual MAS benchmark dataset until now. We construct one as follows. 

\noindent \textbf{Data Source and Data Construction.} Based on the XL-Sum dataset~\cite{hasan-etal-2021-xl}, we construct a \textbf{M}ultilingual \textbf{M}ultimodal abstractive \textbf{Sum}marization (MM-Sum) dataset. The original XL-Sum dataset is crawled from the BBC website\footnote{\url{https://www.bbc.com/}} and its quality has been verified and ensured reliability by~\citet{hasan-etal-2021-xl}. However, the lack of associated image sequence in XL-Sum, makes it impossible to directly conduct research on MAS. Therefore, we strictly follow the procedure of~\cite{hasan-etal-2021-xl} to further offer the image sequence for the corresponding textual summarization dataset, where we maintain the article-summary pair if it contains images and keep the image order appearing in the article. 

\noindent \textbf{Dataset Statistics and Splits.}
\label{dsa}
\autoref{tab:dataset} of~\autoref{data-appendix} shows the detailed statistic of our MM-Sum and please refer to it for details. According to the dataset size of each language, we split them into three settings: Mid-High Resource, Low Resource, and Zero Resource. For mid-high and low-resource languages, following~\citet{hasan-etal-2021-xl}, we utilize about 80\% training:10\% validation:10\% test splitting with one exception (English splitting is 93\%:3.5\%:3.5\%). For zero resource, we following~\citet{DBLP:journals/corr/abs-2201-11732} investigate two scenarios: few-shot and zero-shot. Therefore, we also randomly sample 100 instances as the few-shot learning data and then split the rest with about 50\% validation and 50\% test. 

\subsection{Setup and Metrics}
\label{sect:data}

\noindent \textbf{Implementation Details.}
Please refer to~\autoref{ID} for implementation details including data pre-processing and hyper-parameters settings.

\noindent \textbf{Metrics.}
Following~\citet{hasan-etal-2021-xl}, we use the standard ROUGE scores (R-1, R-2, and R-L)~\cite{lin-2004-rouge} with the statistical significance test~\cite{koehn-2004-statistical} for a fair comparison. 

\begin{table*}[h]
\begin{minipage}{\textwidth}
\centering
\scalebox{0.70}{
\setlength{\tabcolsep}{1.4mm}{
\begin{tabular}{l|c|c|c||c|c|c}
\hline
&\multicolumn{3}{c||}{\textbf{Monolingual Training}} &\multicolumn{3}{c}{\textbf{Multilingual Training}}  \\ 
\cline{2-4} \cline{5-7}
\textbf{Languages} &\textbf{mT5}&\textbf{VG-mT5} &\textbf{SOV-MAS} (ours) &\textbf{mT5}&\textbf{VG-mT5} &\textbf{SOV-MAS} (ours)\\
\hline
Bengali &25.34/\ \;9.52/22.04 &26.02/\ \;9.88/22.14 &\cellcolor{gray!20}{\uwave{26.76/10.08/23.07}} &27.95/10.64/23.43&27.34/10.87/23.42 &\cellcolor{blue!12}{\uline{28.89/11.69/24.59}} \\
French &32.05/12.98/25.06&32.41/13.40/25.50 &\cellcolor{gray!20}{\uwave{33.16/14.21/25.89}} &34.36/14.90/26.92&34.94/15.41/27.56 &\cellcolor{blue!12}{\uline{36.06/16.36/28.63}}  \\
Gujarati &19.30/\ \;6.34/17.74&19.45/\ \;6.26/17.65 &19.83/\ \;6.64/18.02 &21.59/\ \;7.38/19.26&21.44/\ \;7.61/19.46 &\cellcolor{blue!12}{\uline{22.31/\ \;8.12/20.14}} \\
Hausa &36.36/15.37/28.85&35.69/14.75/28.22 &\cellcolor{blue!12}{\uline{36.81/15.31/29.12}} &38.37/16.59/30.34&38.14/16.60/30.45 &\cellcolor{blue!12}{\uline{39.40/17.53/31.04}}  \\
Japanese &44.54/21.33/34.44&45.03/21.64/34.99 &\cellcolor{gray!20}{\uwave{45.97/22.63/35.84}} &47.36/22.20/35.88&46.65/22.66/35.68 &\cellcolor{blue!12}{\uline{47.96/23.76/36.78}}  \\
Marathi &20.39/\ \;8.96/18.65&20.60/\ \;9.06/18.75&21.08/\ \;9.46/19.09 &21.91/\ \;9.52/19.64&21.72/\ \;9.49/19.82 &\cellcolor{blue!12}{\uline{22.59/\ \;9.98/20.39}} \\
Oromo &15.91/\ \;5.03/13.91&15.65/\ \;4.95/13.67 &\cellcolor{blue!12}{\uline{16.68/\ \;5.39/14.60}} &17.77/\ \;5.72/15.53 &17.82/\ \;5.75/15.20   &\cellcolor{blue!12}{\uline{19.13/\ \;6.29/16.47}}  \\
Pashto &36.14/14.06/29.74 &35.97/14.08/29.67 &\cellcolor{gray!20}{\uwave{36.45/14.06/29.79}} &37.34/14.41/30.39&37.21/14.70/30.59  &\cellcolor{blue!12}{\uline{38.11/15.53/31.44}} \\
Pidgin &35.22/12.93/27.27&35.14/12.88/27.27 &35.58/13.02/27.46  &36.33/13.60/28.29&37.21/14.48/29.14 &\cellcolor{blue!12}{\uline{38.02/15.31/30.07}}  \\
Punjabi &27.43/10.07/22.68&27.27/\ \;9.76/22.44  &\cellcolor{gray!20}{\uwave{28.25/10.57/23.14}} &29.98/11.14/24.41&29.75/11.48/24.72 &\cellcolor{blue!12}{\uline{30.78/12.10/25.52}} \\
Serbian Cyrillic &18.52/\ \;4.90/15.44&19.01/\ \;4.92/15.72 &\cellcolor{gray!20}{\uwave{19.80/\ \;5.20/16.41}} &23.11/\ \;7.18/19.14 &22.92/\ \;7.43/19.39 &\cellcolor{blue!12} {\uline{23.85/\ \;7.93/20.06}}  \\
Serbian Latin &18.50/\ \;4.40/15.11&18.49/\ \;4.67/15.42 &18.55/\ \;4.75/15.29 &21.28/\ \;6.04/17.41&20.66/\ \;5.82/17.21 &\cellcolor{blue!12}{\uline{22.39/\ \;6.84/18.59}}  \\
Swahili &34.22/14.76/27.61&34.79/15.07/28.00 &34.56/14.99/27.75 &36.75/16.26/29.49&37.19/17.23/30.33 &\cellcolor{blue!12}{\uline{38.04/17.87/30.99}} \\
Telugu &17.06/\ \;5.83/15.29&17.20/\ \;5.95/15.30 &17.56/\ \;6.09/15.66 &18.68/\ \;6.50/16.52&18.92/\ \;6.77/16.84 &\cellcolor{blue!12}{\uline{20.19/\ \;7.38/17.91}} \\
Welsh &30.41/\ \;9.23/24.11  &30.63/\ \;9.78/24.23 &\cellcolor{gray!20}{\uwave{31.32/10.97/24.77}} &31.86/10.88/25.06 &31.91/10.62/25.08 &\cellcolor{blue!12}{\uline{32.89/11.79/26.10}}  \\

\cline{1-7}
\textbf{Avg.} &27.42/10.38/22.52  &27.55/10.47/22.59 &28.16/10.90/23.06 &29.64/11.53/24.11 &{29.59/11.79/24.32} &\textbf{30.71}/\textbf{12.57}/\textbf{25.25}  \\
\hline
\end{tabular}}}
\caption{The R-1/R-2/R-L results on the low-resource scenario.}\label{tab:Low}
\end{minipage}
\end{table*}

\subsection{Comparison Models}
\label{ssec:layout}
\noindent\textbf{Text-Only MAS Systems.}
\begin{itemize}[leftmargin=*]
\item \textbf{mT5}: We choose the mT5~\cite{xue-etal-2021-mt5}, a multilingual language model pre-trained on a large dataset of 101 languages, as the text-only baseline which is fine-tuned on our dataset.
\end{itemize}
\noindent\textbf{Vision-Guided MAS Systems.}
\begin{itemize}[leftmargin=*]
\item \textbf{VG-mT5}: We implement the fusion method described in~\autoref{MAS} to inject visual features into the mT5 model, which is a strong baseline.
\item \textbf{SOV-MAS}: It is the proposed model with two summary-oriented auxiliary tasks to enhance MAS model as described in~\autoref{sec:sov-mas}.
\end{itemize}

All the above models involve two training manners: \textbf{monolingual training} and \textbf{multilingual training}. Specifically, for~\emph{monolingual training}, we train the model on the training dataset of each language. For~\emph{multilingual training}, we train the model on the whole training dataset of mid-high-resource and low-resource languages.

\subsection{Main Results}

\autoref{tab:Mid-High},~\autoref{tab:Low}, and~\autoref{tab:Zero} present the main results on mid-high-, low-, and zero-resource scenarios under \emph{monolingual} and \emph{multilingual training} settings. Overall, our model obtains notably better results than the text-only ``mT5'' model on both settings. %
1) In the \emph{monolingual training} setting, we find that the fewer the data are (mid-high$\rightarrow$low$\rightarrow$zero), the greater the improvement we gain, showing that our approach plays an increasing role in vision modeling. 2) In the \emph{multilingual training} setting, the results show that our approach learns transferable visual features among languages, especially on the zero-resource ones where the vision serves as an anchor. These results not only show the effectiveness of our approach but also the value of our MM-Sum dataset.

\paragraph{Results on Mid-High-Resource Scenario.} 
\label{ssec:ende}
In \autoref{tab:Mid-High}, 1) on the whole, the results of the \emph{multilingual training} group (\emph{e.g.}, SOV-MAS) substantially outperform those of the \emph{monolingual training} group, demonstrating the task knowledge among languages is transferable. 2) Under the \emph{monolingual training} setting, the text-only baseline ``mT5'' performs worse than the ``VG-mT5'' model on most languages, showing that the visual features indeed supplement some crucial information for the summarization. With the summary-oriented vision modeling tasks, our model further promotes the quality of the summary (``SOV-MAS'' vs. ``VG-mT5''), demonstrating the effectiveness of our approach. 3) Under the \emph{multilingual training} setting, our model consistently and significantly surpasses both the text-only and vision-guided baselines by large margins (\emph{e.g.}, the previous best ``VG-mT5'', up to \textbf{1.20/0.87/1.01} ROUGE scores on average).

Further, in the monolingual setting, the data scale is large while it may be not enough to learn better summary-oriented image features. That's, the improved image features may not supplement much more information compared with the large textual data. However, in multilingual training, the data scale is much larger and enough for learning the better summary-oriented image features, which help the model capture more summary-related information. Thus, the SOV-MAS achieves more significant results than in a monolingual setting.

\paragraph{Results on Low-Resource Scenario.}
\label{ssec:chen}
Under the low-resource languages, in~\autoref{tab:Low}, we observe similar findings as in the Mid-High-Resource scenario. This demonstrates that our conclusions are solid and convincing on general languages. All these results prove the effectiveness of our approach.

Further, in this setting, the data may be not enough for learning the better summary-oriented image features. However, the learned image features still could offer a sketch of the summary and help the model to focus more on the summary-related parts. This may compensate for the impact of insufficient data. Therefore, the SOV-MAS also obtains significant gains.

\begin{table*}[t!]
\centering
\scalebox{0.71}{
\setlength{\tabcolsep}{1.5mm}{
\begin{tabular}{l|c|c|c||c|c|c}
\hline
&\multicolumn{3}{c||}{\textbf{Zero-Shot Setting}} &\multicolumn{3}{c}{\textbf{Few-Shot Setting}}  \\ 
\cline{2-4} \cline{5-7}
\textbf{Languages} &\textbf{mT5}&\textbf{VG-mT5} &\textbf{SOV-MAS} (ours) &\textbf{mT5}&\textbf{VG-mT5} &\textbf{SOV-MAS} (ours) \\
\hline
Amharic &\ \;0.05/0.00/\ \;0.05 &\ \;0.06/0.01/\ \;0.07 &\ \;0.15/0.01/\ \;0.15 &10.50/\ \;2.50/\ \;9.39 &\ \;10.86/\ \;2.58/\ \;9.68 &\ \;9.61/\ \;2.06/\ \;8.33\\
Azerbaijani &\ \;6.79/1.66/\ \;6.25&\ \;6.92/1.76/\ \;6.42 &\cellcolor{gray!20}{\uwave{\ \;7.55/1.93/\ \;6.99}}  &10.57/\ \;2.85/\ \;9.39&\ \;10.91/\ \;3.07/\ \;9.80 &\cellcolor{blue!12}{\uline{12.39/\ \;3.53/10.93}}\\
Burmese &\ \;1.21/0.71/\ \;1.07&\ \;1.27/0.67/\ \;1.11 &\ \;1.41/0.74/\ \;1.18 &33.67/14.16/23.67&33.45/14.23/23.77 &32.97/13.12/22.87 \\
Igbo &18.61/3.00/14.00&19.35/3.61/14.78 &\cellcolor{blue!12}{\uline{21.21/4.08/15.95}}  &21.83/\ \;4.53/16.62&24.17/\ \;5.16/18.14 &24.63/\ \;5.47/18.21 \\
Kirundi &14.39/4.15/11.75&15.70/4.93/13.10 &\cellcolor{blue!12}{\uline{17.31/5.39/14.29}}  &22.09/\ \;6.65/16.81&23.35/\ \;7.28/17.76 &\cellcolor{blue!12}{\uline{24.61/\ \;8.15/18.65}}  \\
Korean &\ \;1.07/0.03/\ \;1.04&\ \;1.23/0.02/\ \;1.23 &\ \;1.13/0.04/\ \;1.09 &\ \;9.49/\ \;4.47/\ \;8.90&10.00/\ \;4.73/\ \;9.41 &\ \;8.65/\ \;4.22/\ \;8.15\\
Kyrgyz &\ \;4.99/1.55/\ \;4.70&\ \;5.52/1.61/\ \;5.19 &\cellcolor{gray!20}{\uwave{\ \;6.40/1.82/\ \;5.85}} &\ \;9.20/\ \;2.25/\ \;7.83&\ \;9.98/\ \;2.67/\ \;8.75 &\cellcolor{gray!20}{\uwave{10.96/\ \;2.96/\ \;9.37}} \\
Nepali &10.62/2.27/\ \;9.53&11.58/2.55/10.10 &\cellcolor{blue!12}{\uline{12.92/3.01/11.42}}  &18.39/\ \;5.24/16.55&18.86/\ \;5.48/17.01 &\cellcolor{blue!12}{\uline{20.11/\ \;6.18/18.11}} \\
Scottish Gaelic &\ \;7.46/0.91/\ \;6.63&\ \;6.61/1.11/\ \;6.01 &\cellcolor{blue!12}{\uline{\ \;8.03/1.45/\ \;7.01}}  &21.68/\ \;5.55/16.96&20.99/\ \;6.32/17.03 &\cellcolor{blue!12}{\uline{24.25/\ \;6.59/18.85}}    \\
Sinhala &\ \;0.11/0.00/\ \;0.11&\ \;0.12/0.01/\ \;0.12  &\ \;0.15/0.01/\ \;0.14  &14.82/\ \;5.28/12.77&14.12/\ \;5.24/12.14  &{{13.76/\ \;4.52/11.48}}\\
Somali &\ \;9.32/1.89/\ \;7.76&\ \;9.58/2.37/\ \;8.13 &\cellcolor{blue!12}{\uline{11.64/2.70/\ \;9.65}} &23.96/\ \;5.43/16.93&23.96/\ \;5.72/17.34 &\cellcolor{blue!12}{\uline{26.26/\ \;6.71/18.79}} \\
Thai &16.34/0.74/16.21&17.79/0.72/17.60 &17.83/0.73/17.67  &24.09/\ \;4.88/18.36&23.76/\ \;4.45/17.65 &\cellcolor{blue!12}{\uline{24.89/\ \;4.42/19.55}}    \\
Tigrinya &\ \;0.08/0.01/\ \;0.08&\ \;0.08/0.01/\ \;0.08 &\ \;0.13/0.00/\ \;0.12 &16.49/\ \;3.35/13.46&16.59/\ \;3.30/13.47 &{{14.50/\ \;2.29/11.84}}   \\
Uzbek &\ \;3.49/0.65/\ \;3.25&\ \;4.77/1.01/\ \;4.46 &\cellcolor{blue!12}{\uline{\ \;6.02/1.32/\ \;5.54}} &\ \;9.83/\ \;2.31/\ \;8.54&10.18/\ \;2.43/\ \;8.98 &\cellcolor{blue!12}{\uline{11.36/\ \;2.96/\ \;9.87}}\\
Yoruba &11.01/2.16/\ \;9.11  &13.38/2.70/10.54 &12.61/2.64/10.18 &24.39/\ \;6.49/18.07  &24.84/\ \;6.58/18.23 &\cellcolor{blue!12}{\uline{26.06/\ \;7.22/19.16}}  \\
\cline{1-7}
\textbf{Avg.} &\ \;7.03/1.31/\ \;6.10  &\ \;7.59/1.53/\ \;6.59 &\ \;\textbf{8.30}/\textbf{1.72}/\ \;\textbf{7.15} &18.07/\ \;5.07/14.29  &18.40/\ \;5.28/14.61 &\textbf{19.00}/\ \;\textbf{5.36}/\textbf{14.96}   \\
\hline
\end{tabular}}}
\caption{The R-1/R-2/R-L results on the zero-resource scenario, which includes zero-shot and few-shot settings.}\label{tab:Zero}
\end{table*}

\paragraph{Results on Zero-Resource Scenario (Zero-Shot).}
On the zero-shot setting in the left group of~\autoref{tab:Zero}, the ``VG-mT5'' model notably exceeds the text-only ``mT5'' model by averagely 0.56/0.22/0.49$\uparrow$ ROUGE scores. It indicates that the image in our MM-Sum plays a key role when transferring knowledge from mid-high and low-resource languages to zero-resource languages via considering vision as the anchor, where the vision is free from different languages. Furthermore, our model presents significant improvements over the ``mT5'' model by averagely \textbf{1.27/0.41/1.05}$\uparrow$ ROUGE gains, which shows its effectiveness again.

\paragraph{Results on Zero-Resource Scenario (Few-Shot).} On the few-shot setting, we merge the 100 samples of each zero-resource language to continue training the \emph{multilingual training} model for 3,000 steps. The results are shown in the right group of~\autoref{tab:Zero}, which shows that with a handful of data the models can greatly increase the ROUGE scores compared with zero-shot results. Our approach still achieves the best results, showing the effectiveness of our approach again. It also suggests that there is much room for further improvement using more data or other more advanced text-vision fusion methods.

Besides, we listed the results with the visual features extracted by the pretrained Transformer vision encoder, \emph{i.e.}, ViT~\cite{dosovitskiy2020image}, in~\autoref{tab:Mid-High-vit} and~\autoref{tab:Low-vit} of the appendix, demonstrating that our SOV-MAS still achieves better performance in almost all cases, showing its superiority.

\section{Analysis}
\subsection{Ablation Study}
\label{ssec:abs}
We conduct ablation studies to investigate how well the two auxiliary tasks work. The results are shown in~\autoref{tab:abl}. We have the following findings: 

\begin{itemize}[leftmargin=*]
\item The Vis2Sum task shows a positive impact on the model performance (row 1 vs. row 0), demonstrating that the image sequence may reflect a sketch of the summary, which is beneficial to the summary generation;
\item The MIM substantially improves the MAS model in terms of ROUGE scores (row 2 vs. row 0), suggesting that reconstructing the masked image with the summary is helpful to summarization; 
\item  The two summary-oriented vision modeling tasks exhibit notable cumulative benefits (row 3 vs. rows 0$\sim$2), showing that focusing on the summary-oriented visual features is effective; 
\item The variant MRM makes relatively smaller contributions to the MAS model compared with the MIM (row 4 vs. row 2). The reason may be that it is easy for the concise summary to complete the masked globally full image rather than the masked locally disordered regions (actually, the local regions might not be mentioned in the summary as described in \autoref{intro}, and thus it is hard to reconstruct them given the concise summary).
\end{itemize}



\begin{table}[t!]
\centering
\scalebox{0.62}{
\setlength{\tabcolsep}{0.50mm}{
\begin{tabular}{l|l|c|c|c}
\hline
&\textbf{Models} &\textbf{Mid-High Resource}&\textbf{Low Resource} &\textbf{Zero Resource} \\
\hline
0&Baseline &32.84/14.49/26.82  &29.59/11.79/24.32 &7.59/1.53/6.59 \\
1&\emph{w/} Vis2Sum &33.74/15.12/27.56 &30.43/12.37/25.01&8.16/1.68/7.07 \\
2&\emph{w/} MIM &33.59/15.04/27.48 &30.37/12.21/24.94&7.93/1.65/6.98  \\
3&\emph{w/} Vis2Sum\&MIM &34.04/15.36/27.83 &30.71/12.57/25.25&8.30/1.72/7.15  \\
\cdashline{1-5}[4pt/2pt]
4&\emph{w/} MRM &33.18/14.58/26.92 &29.99/11.85/24.43&7.68/1.57/6.65  \\
\hline
\end{tabular}}}
\caption{Ablation results under the \emph{multilingual training} setting (Avg. R-1/R-2/R-L results), where each auxiliary task is separately added on the baseline.}\label{tab:abl}
\end{table}

\subsection{Human Evaluation}
To further evaluate the performances of mT5, VG-mT5 and our SOV-MAS, we conduct human studies on 50 samples randomly selected from English and Chinese test sets. We invited three Chinese postgraduate students who are highly proficient in English comprehension~\footnote{One student has passed TEM-8 (with 81 points out of 100 points). The other two students have passed the IELTS exam (their scores of reading comprehension are 8.0 and 7.0 out of 9.0 points, respectively)} to compare the generated summaries under the multilingual training setting and assess each summary from three independent perspectives: \textbf{fluency} (Flu.), \textbf{conciseness} (Conci.) and \textbf{informativeness} (Info.). We ask them to assess each aspect with a score ranging from 1 (worst) to 5 (best). The average results are presented in~\autoref{human1}. 

\autoref{human1} shows the human results on English and Chinese. We find that our SOV-MAS outperforms all compared models from all criteria in both languages, which further demonstrates the effectiveness and superiority of our model. The Fleiss' Kappa scores~\cite{doi:10.1177/001316447303300309} of Flu., Conci and Info. are 0.69, 0.65 and 0.56, respectively, which indicates a substantial agreement among three evaluators. We also present a case study in~\autoref{CS}.

\begin{table}[]
\small
\centering
\scalebox{0.9}{
\begin{tabular}{@{}l|ccc|ccc@{}}
\hline
\multirow{2}{*}{\textbf{Models}} & \multicolumn{3}{c|}{\textbf{English}} & \multicolumn{3}{c}{\textbf{Chinese}} \\ \cline{2-7} 
 & Flu. & Conci. & Info. & Flu. & Conci. & Info. \\ 
\hline
mT5 & 4.04 & 3.86 & 3.18 & 3.42 & 3.20 & 3.08 \\
VG-mT5 & 4.22 & 4.08 & 3.36 & 3.74 & 3.42 & 3.26 \\
SOV-MAS & \textbf{4.56} & \textbf{4.38} & \cellcolor{gray!20}{\uwave{\textbf{3.88}}} & \textbf{3.98} & \textbf{3.76} & \cellcolor{gray!20}{\uwave{\textbf{3.64}}} \\ 
\hline
\end{tabular}}
\caption{Human evaluation results in terms of fluency (Flu.), conciseness (Conci.) and informativeness (Info.).}
\label{human1}
\end{table}

\subsection{Results on How2 Dataset}
To investigate the generality of the two summary-oriented vision modeling tasks, we extend them to two existing MAS models (\emph{i.e.}, VG-T5 and VG-BART~\cite{yu-etal-2021-vision}), denoted as ``SOV-MAS (T5)'' and ``SOV-MAS (BART)'', respectively. As shown in~\autoref{tbl:w_f}, we also compare our models with the following systems, including text-only models: S2S, PG, Trans., T5, and BART, and prior best vision-guided models: HA (RNN/Trans.), MFFG (RNN/Trans.), VG-T5, and VG-BART. 

The results on How2 dataset~\cite{sanabria18how2}, a widely-used English MAS dataset, show that our approach effectively boosts the model performance and notably outperforms both text-only and vision-guided methods, suggesting the effectiveness and generalizability of our approach.

\section{Related Work}

\paragraph{Abstractive Text Summarization (ATS).} 
Given the input textual article, the goal of ATS is to generate a concise summary~\cite{10.5555/2969239.2969428,https://doi.org/10.48550/arxiv.2203.12515}. Thanks to generative pre-trained language models~\cite{lewis-etal-2020-bart}, ATS has achieved remarkable performance~\cite{paulus2018a,liu-lapata-2019-text,pmlr-v119-zhang20ae,goodwin-etal-2020-flight,rothe-etal-2021-thorough,xiao-etal-2022-primera,xu-etal-2020-self,yu-etal-2021-adaptsum,liang-etal-2022-variational,wang2022understanding}.

\begin{table}[t!]
\centering
\scalebox{0.80}{
\setlength{\tabcolsep}{0.9mm}{
\begin{tabular}{c|l|c}
\hline
\multirow{5}{*}{{T}}
&{S2S}~\cite{luong-etal-2015-effective}$^*$ &58.6/40.6/53.8          \\
&{PG}~\cite{DBLP:journals/corr/SeeLM17}$^*$   &57.2/39.5/52.8 \\
&{Transf.}~\cite{vaswani2017attention}$^*$  &59.0/41.0/54.3  \\ 
&{T5}~\cite{2020t5}$^*$   &62.8/45.0/57.5 \\
&{BART}~\cite{lewis-etal-2020-bart}$^*$   &64.0/46.4/58.9 \\
\hline
\multirow{8}{*}{{T+V}}
&{HA (RNN)}~\cite{palaskar-etal-2019-multimodal}$^*$    &60.3/42.5/55.7  \\
&{HA (Trans.)}~\cite{palaskar-etal-2019-multimodal}$^*$      &60.2/43.1/55.9 \\
&{MFFG (RNN)}~\cite{liu-etal-2020-multistage}$^*$   &62.3/46.1/58.2 \\
&{MFFG (Trans.)}~\cite{liu-etal-2020-multistage}$^*$   &61.6/45.1/57.4 \\ 
&VG-T5~\cite{yu-etal-2021-vision}$^{*\dagger}$    &63.3/45.3/58.0 \\
&VG-BART~\cite{yu-etal-2021-vision}$^{*\dagger}$    &66.3/49.4/61.4 \\
\cdashline{2-3}[4pt/2pt]
    &{SOV-MAS} (T5)  &\cellcolor{blue!12}{\uline {64.8/46.7/59.5}}  \\ 
&{SOV-MAS} (BART)  &\cellcolor{blue!12}{\uline {\textbf{67.7}/\textbf{50.9}/\textbf{62.8}}}\\ 
\hline
\end{tabular}}}
\caption{The R-1/R-2/R-L results on test sets of How2 dataset~\cite{sanabria18how2}. ``$^*$'' indicates that the results are taken from~\citet{yu-etal-2021-vision}. ``$^\dagger$'' indicates the previous state-of-the-art models. T/V: text/vision.}
\label{tbl:w_f}
\end{table}

\paragraph{Multimodal Abstractive Summarization (MAS).} 
With the rapid growth of multimedia, many MAS datasets have been built such as: SportsSum~\cite{5711541}, MovieSum~\cite{6527322}, MSMR~\cite{1221239}, MMSS~\cite{li-etal-2017-multi}, MSS~\cite{li2018multi}, How2~\cite{sanabria18how2}, MSMO~\cite{zhu-etal-2018-msmo}, E-DailyMail~\cite{chen-zhuge-2018-abstractive}, EC-product~\cite{li2020aspect}, and MM-AVS~\cite{fu-etal-2021-mm}. All these datasets, covering video summarization, movie summarization, meeting records summarization, sentence summarization, product summarization, and news summarization, aim to generate a summary based on multimodal inputs (text, vision, or audio). With the data resources extensively used, the MAS task has attracted much attention, where the existing work mainly focuses on how to effectively exploit the additional features which are generally implicitly learned by the MAS objective, having achieved impressive performance on these high-resource English datasets~\cite{li2018read,li-etal-2020-vmsmo,zhu2020multimodal,zhu2021graph,zhang2021unims,zhang2021hierarchical,yu-etal-2021-vision}. For example, ~\citet{palaskar-etal-2019-multimodal} and~\citet{zhang2021hierarchical} explore the hierarchy between the textual article and visual features, and integrate them into the MAS model.~\citet{liu-etal-2020-multistage} design a multistage fusion network to model the fine-grained interactions between the two modalities. And~\citet{yu-etal-2021-vision} study multiple multimodal fusion methods to infuse the visual features into generative pre-trained language models, \emph{e.g.}, BART~\cite{lewis-etal-2020-bart}.

\paragraph{Multilingual Abstractive Summarization.} 
It aims to train a model that can produce a summary in any language. Existing studies mainly pay attention to constructing the multilingual abstractive summarization dataset and there have been many datasets publicly available: MultiLing2015~\cite{giannakopoulos-etal-2015-multiling}, GlobalVoices~\cite{nguyen-daume-iii-2019-global}, MultiSumm~\cite{Cao_Wan_Yao_Yu_2020}, MLSUM~\cite{scialom-etal-2020-mlsum}, MultiHumES~\cite{yela-bello-etal-2021-multihumes}, MassiveSumm~\cite{varab-schluter-2021-massivesumm}, MLGSum~\cite{wang-etal-2021-contrastive}, and XL-Sum~\cite{hasan-etal-2021-xl}. Most of these datasets are automatically constructed from online websites due to high human cost, which involves at least two languages. 

There are two essential differences between the above work and ours:

$\romannumeral1$) The MAS datasets and multilingual abstractive summarization datasets are either in multimodal or multilingual, while ours includes both. It is obvious that conducting multilingual MAS is more challenging due to the more complex scene~\cite{DBLP:journals/corr/abs-2109-05199}. Besides, our MM-Sum includes 44 languages, covering three settings: mid-high, low, and zero resource. What is more, our MM-Sum has the property that the knowledge can be transferred from mid-high resource languages to low- and zero-resource ones through visual features (as the bridge) while they have not.~\autoref{data:related} of~\autoref{DC} provides a detailed comparison of available languages, modalities, and scenes for all datasets.

$\romannumeral2$) We mainly focus on how to obtain the summary-oriented visual features from the perspective of the summary rather than the article as existing work does. We thus propose two summary-oriented vision modeling tasks which are flexible and easy to be extended to existing MAS models.

\section{Conclusion}
In this paper, we propose to enhance the MAS model through two summary-oriented vision modeling tasks namely \emph{vision to summary task} and \emph{masked image modeling task}. They can explicitly force the MAS model to exploit the summary-oriented visual features and thus improve the summary quality. Extensive experiments on multiple settings demonstrate that our model significantly outperforms related baselines in terms of ROUGE scores and human evaluation. Furthermore, we contribute a large-scale multilingual MAS (MM-Sum) dataset to the research community. 

\section*{Limitations}

Although we show that our SOV-MAS outperforms the VG-mT5 model under different setups, there are some limitations worth considering to study in future work: (1) In this study, we only provide 44 languages and conduct experiments on them, and future work could extend our method to more languages; (2) The used MAS model is based on the generative pre-trained language model, \emph{i.e.}, mT5~\cite{xue-etal-2021-mt5}. The large-scale model size can bring promising performance while it also consumes more training time (all mT5-based models in this work cost about five days under the multilingual training setting) and releases more carbon dioxide, which may be inconsistent with the theme of green AI. Therefore, the work related to model compression (\emph{e.g.}, knowledge distillation) may be possibly future work for the multilingual MAS task. 

\section*{Ethics Statement}
In this section, we consider the potential ethical issues of our model. In this paper, we propose SOV-MAS which is trained on the publicly-available BBC datasets. Therefore, SOV-MAS might lead to incorrect summaries in applications and involve the same biases and toxic behaviors exhibited by the datasets. Besides, we crawled the dataset from the BBC website\footnote{https://www.bbc.com/} and its permissions are granted to copy, distribute and modify the contents under the terms of the \href{https://en.wikipedia.org/wiki/Wikipedia:Text_of_Creative_Commons_Attribution-ShareAlike_3.0_Unported_License}{Creative Commons AttributionShareAlike 3.0 Unported License} and \href{https://www.wikidata.org/wiki/Wikidata:Text_of_the_Creative_Commons_Public_Domain_Dedication}{Creative Commons CC0 License}, respectively.

\section*{Acknowledgements}
The research work described in this paper has been supported by the National Key R\&D Program of China (2020AAA0108001) and the National Nature Science Foundation of China (No. 61976015, 61976016, 61876198 and  61370130). The authors would like to thank the anonymous reviewers for their insightful comments and suggestions to improve this paper.

\bibliography{anthology,custom}
\bibliographystyle{acl_natbib}

\clearpage
\newpage
\appendix
\section{Dataset Statistics and Splits.}
\label{data-appendix}
\autoref{tab:dataset} shows that our MM-Sum covers 44 languages and in total includes 1,078,215 article-summary pairs with 3,479,348 images, where each article-summary pair contains about 3.23 images on average. The average article and summary length for all languages are about 520 and 84, respectively. According to the dataset size of each language, we split them into three settings: Mid-High Resource, Low Resource, and Zero Resource. For mid-high and low-resource languages, following~\citet{hasan-etal-2021-xl}, we utilize about 80\% training:10\% validation:10\% test splitting with one exception (English splitting is 93\%:3.5\%:3.5\%). For zero resource, we follow~\citet{DBLP:journals/corr/abs-2201-11732} who investigate two scenarios: few-shot and zero-shot. Therefore, we also randomly sample 100 instances as the few-shot learning data and then split the rest with about 50\% validation and 50\% test.

\begin{table*}[h]
\begin{minipage}{\textwidth}
\centering
\scalebox{0.78}{
\setlength{\tabcolsep}{1.50mm}{
\begin{tabular}{lrr|lrr|lrr}
\hline
\multicolumn{3}{c|}{\textbf{Mid-High Resource}} &\multicolumn{3}{c|}{\textbf{Low Resource}}    &\multicolumn{3}{c}{\textbf{Zero Resource}} \\ 
\textbf{Languages} &\textbf{\#Samples}&\textbf{\#Images} &\textbf{Languages} &\textbf{\#Samples}&\textbf{\#Images} &\textbf{Languages} &\textbf{\#Samples}&\textbf{\#Images}\\
\hline
Arabic &41,977&95,762 &Bengali &10,008&33,447 &Amharic &7,153&11,895\\
Chinese &41,126&101,672 &French &10,478&23,698 &Azerbaijani &7,392&21,612 \\
English &311,999&867,817 &Gujarati &10,917&72,196 &Burmese &5,614&13,727 \\
Hindi &49,059&209,559 &Hausa &7,536&17,023 &Igbo &4,773&17,113 \\
Indonesian &45,248&132,048 &Japanese &8,802&25,261 &Korean &5,049&15,908 \\
Persian &29,547&87,768 &Marathi &12,354&59,553 &Kyrgyz &3,187&11,169 \\
Portuguese &25,230&124,136 &Oromo &7,551&16,160   &Kirundi &7,088&15,352 \\
Russian &65,276&216,237 &Pashto &15,683&33,851  &Nepali &6,766&18,891 \\
Spanish &45,730&219,365 &Pidgin &11,173&26,031 &Scottish Gaelic &2,303&14,213 \\
Tamil &19,939&72,441  &Punjabi &10,068&46,874 &Sinhala &3,192&8,198 \\
Turkish &21,970&61,443 &Serbian Cyrillic &8,737&39,577 &Somali &7,358&17,545 \\
Ukrainian &34,202&117,587 &Serbian Latin &8,737&39,561 &Tigrinya &6,790&14,777 \\
Urdu &40,672&106,960 &Swahili &9,825&26,770 &Thai &7,339&31,414 \\
Vietnamese &23,100&62,436 &Telugu &12,388&58,206 &Uzbek &4,421&11,840 \\
\cline{1-3}
\textbf{Total Samples} &\textbf{1,078,215} &&Welsh &12,162&140,638 &Yoruba &7,368&20,388 \\
\cline{4-9}
\textbf{Total Images}  &\textbf{3,479,348} && \textbf{Avg. of Images} &\textbf{3.23}&&\textbf{Num. of Lang.} &\textbf{44} &\\
\hline
\end{tabular}}}
\caption{Languages covered by our MM-Sum dataset, and the number of samples with corresponding images for each language. Here, a sample denotes an article-summary pair. We roughly split them into three scenarios according to the number of samples, \emph{i.e.}, Mid-High Resource, Low Resource, and Zero Resource.}\label{tab:dataset}
\end{minipage}
\end{table*}

\section{Implementation Details}
\label{ID}
\paragraph{Data Pre-Processing.} Following~\citet{hasan-etal-2021-xl}, we pre-process the textual data by truncating or padding them into sequences of 512 tokens for $\mathcal{X}$ and the outputs $\mathcal{Y}$ to 84 tokens after using the 250k wordpiece~\cite{xue-etal-2021-mt5} vocabulary provided with the mT5 checkpoint. For the image sequence, after the feature extraction as described in~\autoref{sec:vision_features}, we also truncate or pad the sequence length to 180 (\emph{i.e.}, five images: 5 * 36; n=5, m=36).

\paragraph{Hyper-Parameters.} Following~\citet{hasan-etal-2021-xl}, we use the $base$\footnote{\url{https://huggingface.co/google/mt5-base/tree/main}} model of mT5~\cite{xue-etal-2021-mt5}, in which $L$ = 12 for both encoder and decoder. For the vision-related hyper-parameters mentioned in~\autoref{MAS}, we follow~\citet{yu-etal-2021-vision} for a fair comparison. Specifically, we use a 4-layer encoder (\emph{i.e.}, $H$ = 4) with 8 attention heads and a 2048 feed-forward dimension. For all models, the dropout is set to 0.1 and the label smoothing is set to 0.1. The $d$, $d_c$, and $d_v$ are 768, 256, and 2048, respectively. The balancing factor $\alpha$ and $\beta$ in~\autoref{loss_all} are set to 1.0, which are not tuned. The $K$ of~\autoref{loss_all_multi} is 29, which is the sum of the number of mid-high- and low-resource languages. During the \emph{monolingual training}, we train all models on each language separately for 6-20 epochs (since the total training samples were limited, we had to be careful to prevent overfitting) on an NVIDIA Tesla V100 GPU with a batch size of 32. The models are optimized using Adam~\cite{kingma2017adam} with $\beta_1$=0.9 and $\beta_2$=0.998. We train all model weights with a slanted learning rate schedule (learning rate to 5e-4). 
During the \emph{multilingual training}, following a similar training strategy~\cite{10.5555/3454287.3454921,hasan-etal-2021-xl}, we sample each batch from a single language containing 256 samples and use a smoothing factor (0.5) so that batches of low-resource languages would be sampled at a higher rate, increasing their frequency during training. We set the training step to 35,000 steps on a distributed cluster of 8 NVIDIA Tesla V100 GPUs and trained about 5 days. We use the Adafactor optimizer~\cite{pmlr-v80-shazeer18a} with a linear warm-up of 5,000 steps and the ``inverse square root'' learning rate schedule. 

For inference, we use beam search with beam size 4 and length penalty of $\gamma$ = 0.6. When calculating the ROUGE scores, we use the multi-lingual rouge\footnote{\url{https://github.com/csebuetnlp/xl-sum/tree/master/multilingual_rouge_scoring}} toolkit following~\citet{hasan-etal-2021-xl}. All experimental results reported in this paper are the average of three runs with different random seeds. 

\begin{table*}[h]
\begin{minipage}{\textwidth}
\centering
\scalebox{0.72}{
\setlength{\tabcolsep}{1.8mm}{
\begin{tabular}{l|c|c|c||c|c|c}
\hline
&\multicolumn{3}{c||}{\textbf{Monolingual Training}} &\multicolumn{3}{c}{\textbf{Multilingual Training}}  \\ 
\cline{2-4} \cline{5-7}
\textbf{Languages} &\textbf{mT5}&\textbf{VG-mT5} &\textbf{SOV-MAS} (ours) &\textbf{mT5}&\textbf{VG-mT5} &\textbf{SOV-MAS} (ours)\\
\hline
Arabic &33.67/14.06/27.83 &33.79/14.11/27.95 &33.86/14.53/28.06 &34.34/14.30/28.43&33.40/13.49/27.51 &\cellcolor{blue!12}{\uline{34.69/14.39/28.54}} \\
Chinese &40.20/25.39/33.49&40.31/25.45/33.51 &40.61/25.37/33.39 &40.30/24.97/33.04&40.19/25.31/33.35 &\cellcolor{blue!12}{\uline{41.51/26.34/34.41}}  \\
English &36.99/15.18/29.64 &37.25/14.97/29.54 &37.29/15.18/29.82 &36.65/13.91/28.53&36.69/14.16/28.79 &\cellcolor{blue!12}{\uline{37.77/15.14/29.81}} \\
Hindi &33.66/13.14/27.71&34.55/13.47/28.26 &34.78/13.55/28.11 &35.50/13.91/28.52&35.66/14.26/28.97 &\cellcolor{blue!12}{\uline{36.33/14.91/29.68}}  \\
Indonesian &35.10/15.44/28.91&35.16/15.49/29.09 &35.14/15.31/28.81 &35.84/15.66/29.40&36.55/16.38/30.19 &\cellcolor{blue!12}{\uline{37.46/17.13/31.18}}  \\
Persian &36.14/15.55/29.25&36.01/15.45/29.08 &36.37/15.75/29.35 &36.39/15.84/29.45&36.88/16.34/29.93 &\cellcolor{blue!12}{\uline{37.65/16.92/30.58}} \\
Portuguese &30.13/10.32/22.06&29.46/\ \;9.72/21.91 &29.77/10.01/21.55 &30.84/10.92/22.64&31.01/11.22/23.11   &\cellcolor{blue!12}{\uline{31.77/11.76/23.79}}  \\
Russian &30.01/12.47/24.28&31.01/12.43/24.52 &\cellcolor{gray!20}{\uwave{31.58/12.77/24.96}} &31.12/12.33/24.67&30.55/12.65/24.58  &\cellcolor{blue!12}{\uline{31.57/13.12/25.21}} \\
Spanish &29.51/10.48/22.51&29.37/10.59/22.52 &29.19/10.32/22.37 &29.91/10.70/22.66&30.37/10.94/23.02 &\cellcolor{blue!12}{\uline{31.00/11.56/23.58}}  \\
Tamil &22.31/10.08/20.36&22.29/10.14/20.38  &\cellcolor{gray!20}  {\uwave{22.80/10.51/20.62}} &22.96/10.05/20.75&23.14/10.29/20.98 &\cellcolor{blue!12}{\uline{24.01/10.82/21.89}}  \\
Turkish &30.37/14.39/26.79&30.44/14.40/26.77 &\cellcolor{gray!20}  {\uwave{30.91/14.60/27.16}} &31.93/14.69/27.76&31.41/14.71/27.70 &\cellcolor{blue!12}{\uline{32.67/15.70/28.77}}  \\
Ukrainian &21.57/\ \;8.66/18.64&21.69/\ \;8.78/18.65 &21.77/\ \;8.61/18.77 &22.79/\ \;9.13/19.46&22.79/\ \;9.39/19.75 &\cellcolor{blue!12}{\uline{23.84/\ \;9.94/20.49}}  \\
Urdu &38.22/17.25/31.37 &38.11/17.27/31.51 &38.19/17.12/31.38 &38.15/17.12/31.36&38.01/17.21/31.55 &\cellcolor{blue!12}{\uline{39.22/18.31/32.62}} \\
Vietnamese &32.18/15.84/24.83&32.19/15.99/24.87 &\cellcolor{gray!20}{\uwave{32.87/16.59/25.24}} &33.71/16.72/25.97&33.79/17.08/26.34 &\cellcolor{blue!12}{\uline{34.75/17.82/27.09}} \\
\cline{1-7}
\textbf{Avg.} &32.14/14.16/26.26  &32.25/14.16/26.32 &32.49/14.26/26.40 &32.88/14.30/26.61 &32.89/14.53/26.84&\textbf{33.87}/\textbf{15.27}/\textbf{27.69}  \\

\hline
\end{tabular}}}
\caption{The R-1/R-2/R-L results on the mid-high-resource scenario with visual features extracted by Vision Transformer (ViT)~\cite{dosovitskiy2020image}. ``\colorbox{gray!20}{\uwave{*/*/*}}'' and ``\colorbox{blue!12}{\uline{*/*/*}}'' denote statistically significant better than the ``VG-mT5'' model with t-test \emph{p} \textless \ 0.05 and \emph{p} \textless \ 0.01 hereinafter, respectively. The ``Avg.'' indicates the average score for each group and the best average scores are \textbf{bold}.}\label{tab:Mid-High-vit}
\end{minipage}
\end{table*}

\begin{table*}[h]
\begin{minipage}{\textwidth}
\centering
\scalebox{0.72}{
\setlength{\tabcolsep}{1.4mm}{
\begin{tabular}{l|c|c|c||c|c|c}
\hline
&\multicolumn{3}{c||}{\textbf{Monolingual Training}} &\multicolumn{3}{c}{\textbf{Multilingual Training}}  \\ 
\cline{2-4} \cline{5-7}
\textbf{Languages} &\textbf{mT5}&\textbf{VG-mT5} &\textbf{SOV-MAS} (ours) &\textbf{mT5}&\textbf{VG-mT5} &\textbf{SOV-MAS} (ours)\\
\hline
Bengali &25.34/\ \;9.52/22.04 &25.86/\ \;9.81/22.11 &\cellcolor{gray!20}{\uwave{26.49/10.02/23.01}} &27.95/10.64/23.43&27.88/10.82/23.67 &\cellcolor{gray!20}{\uline{28.58/11.45/24.27}} \\
French &32.05/12.98/25.06&32.36/13.35/25.48 &\cellcolor{blue!12}{\uline{33.12/14.21/25.81}} &34.36/14.90/26.92&34.89/15.35/27.39 &\cellcolor{blue!12}{\uline{35.93/16.31/28.42}}  \\
Gujarati &19.30/\ \;6.34/17.74&19.48/\ \;6.29/17.73 &19.81/\ \;6.61/17.89 &21.59/\ \;7.38/19.26&21.49/\ \;7.68/19.47 &\cellcolor{gray!20}{\uline{22.18/\ \;8.21/20.04}} \\
Hausa &36.36/15.37/28.85&35.77/14.88/28.34 &\cellcolor{gray!20}{\uline{36.55/15.12/29.03}} &38.37/16.59/30.34&38.11/16.64/30.47 &\cellcolor{blue!12}{\uline{39.28/17.51/31.01}}  \\
Japanese &44.54/21.33/34.44&44.89/21.62/34.87 &\cellcolor{gray!20}{\uwave{45.91/22.59/35.81}} &47.36/22.20/35.88&46.77/22.61/35.79 &\cellcolor{blue!12}{\uline{47.79/23.67/36.72}}  \\
Marathi &20.39/\ \;8.96/18.65&20.61/\ \;9.09/18.88&21.09/\ \;9.55/19.27 &21.91/\ \;9.52/19.64&21.79/\ \;9.55/19.83 &\cellcolor{blue!12}{\uline{22.61/\ 10.12/20.45}} \\
Oromo &15.91/\ \;5.03/13.91&15.49/\ \;4.95/13.51 &\cellcolor{blue!12}{\uline{16.52/\ \;5.42/14.57}} &17.77/\ \;5.72/15.53 &17.79/\ \;5.79/15.43   &\cellcolor{blue!12}{\uline{18.82/\ \;6.36/16.48}}  \\
Pashto &36.14/14.06/29.74 &36.09/14.10/29.81 &36.41/14.00/29.71 &37.34/14.41/30.39&37.28/14.73/30.63  &\cellcolor{blue!12}{\uline{38.15/15.56/31.46}} \\
Pidgin &35.22/12.93/27.27&35.01/12.67/27.19 &\cellcolor{gray!20}\uwave{{35.59/13.01/27.49}}  &36.33/13.60/28.29&36.88/14.27/29.00 &\cellcolor{blue!12}{\uline{37.91/15.30/30.01}}  \\
Punjabi &27.43/10.07/22.68&27.29/\ \;9.78/22.51  &\cellcolor{blue!12}{\uline{28.27/10.56/23.11}} &29.98/11.14/24.41&29.67/11.35/24.57 &\cellcolor{blue!12}{\uline{30.57/12.02/25.41}} \\
Serbian Cyrillic &18.52/\ \;4.90/15.44&18.96/\ \;4.96/15.75 &\cellcolor{gray!20}{\uwave{19.67/\ \;5.18/16.40}} &23.11/\ \;7.18/19.14 &22.91/\ \;7.41/19.34 &\cellcolor{blue!12} {\uline{23.88/\ \;7.98/20.00}}  \\
Serbian Latin &18.50/\ \;4.40/15.11&18.55/\ \;4.69/15.53 &18.58/\ \;4.88/15.42 &21.28/\ \;6.04/17.41&20.54/\ \;5.80/17.20 &\cellcolor{blue!12}{\uline{21.89/\ \;6.81/18.32}}  \\
Swahili &34.22/14.76/27.61&34.71/15.00/27.91 &34.57/14.95/27.72 &36.75/16.26/29.49&37.13/17.20/30.07 &\cellcolor{blue!12}{\uline{38.02/17.81/30.91}} \\
Telugu &17.06/\ \;5.83/15.29&17.21/\ \;5.98/15.35 &17.51/\ \;6.01/15.61 &18.68/\ \;6.50/16.52&18.93/\ \;6.71/16.80 &\cellcolor{blue!12}{\uline{19.87/\ \;7.33/17.83}} \\
Welsh &30.41/\ \;9.23/24.11  &30.75/\ \;9.73/24.29 &\cellcolor{gray!20}{\uwave{31.31/10.65/24.76}} &31.86/10.88/25.06 &31.90/10.77/25.11 &\cellcolor{blue!12}{\uline{32.86/11.75/26.02}}  \\
\cline{1-7}
\textbf{Avg.} &27.42/10.38/22.52  &27.53/10.452/2.61 &28.09/10.85/23.04 &29.64/11.53/24.11 &29.59/11.77/24.31 &\textbf{30.55}/\textbf{12.54}/\textbf{25.15}  \\
\hline
\end{tabular}}} 
\caption{The R-1/R-2/R-L results on the low-resource scenario with visual features extracted by Vision Transformer (ViT)~\cite{dosovitskiy2020image}.}\label{tab:Low-vit}
\end{minipage}
\end{table*}

\section{Case Study}
\label{CS}
\autoref{cs} shows an example multimodal English document, the generated summary, and the ground truth summary. Though all generated summaries exhibit the core idea of the document and present factual consistency, ours has good lexical and semantics overlaps with the ground truth. And it is not difficult to find that with enhanced visual features our SOV-MAS can capture a sketch of the document, \emph{i.e., mourning the king with true devotion}, and supplement a lot of details, \emph{i.e., dressed in black and weeping}. These observations show that through two summary-oriented vision modeling tasks, our model could generate a better summary. We also believe that a more informative summary would meet the demand of the user.

\textbf{\begin{figure*}[t]
    \centering
    \includegraphics[width=0.98\textwidth]{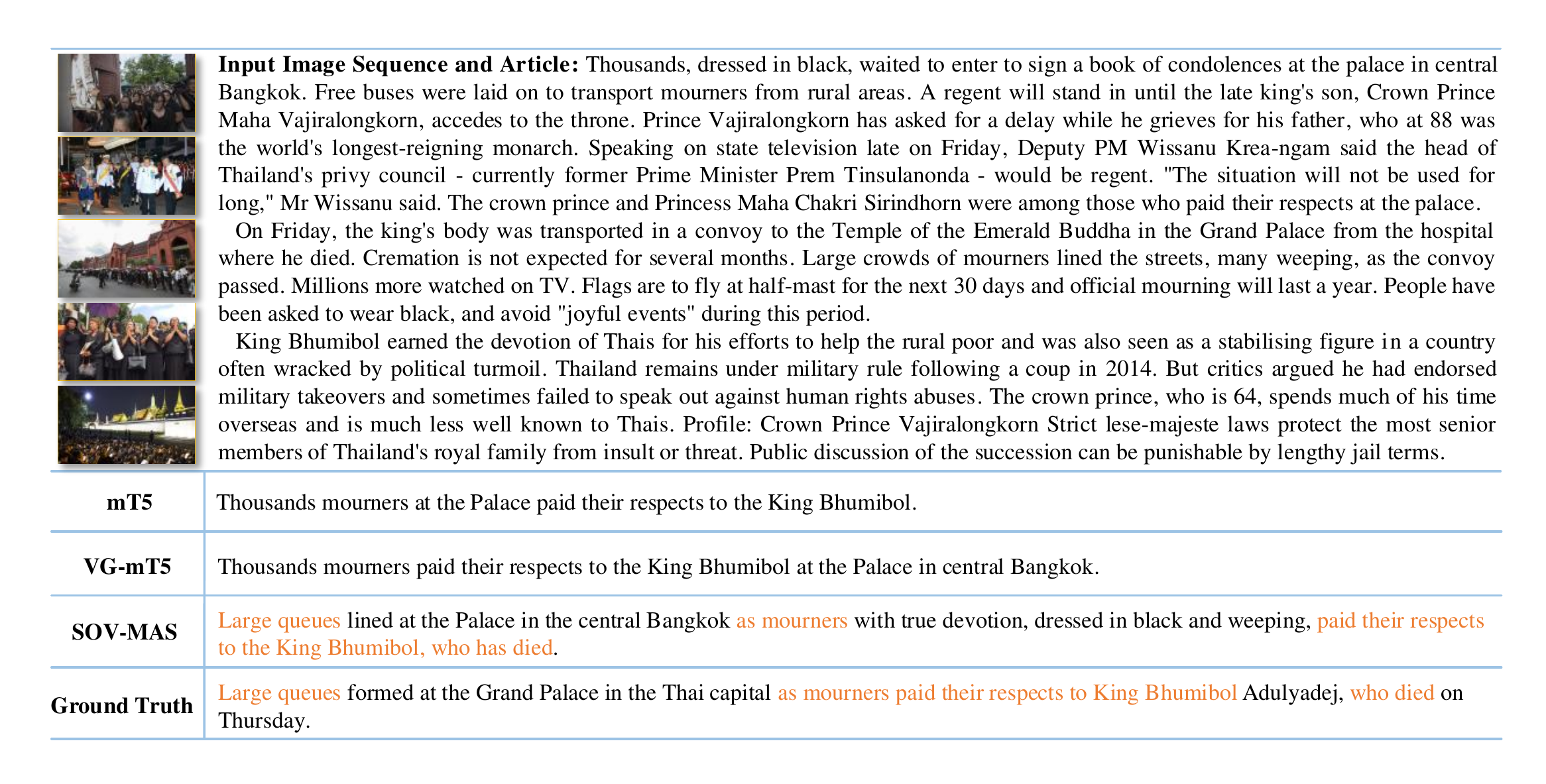}
    \caption{An example of multimodal abstractive summarization in English. }
    \label{cs}
\end{figure*}}

\section{Comparison to the Related Datasets}
\label{DC}
\autoref{data:related} provides information on the number of available languages, modalities, and scenes for all datasets. Specifically, multimodal abstractive summarization datasets and multilingual abstractive datasets are either multimodal or multilingual, while ours includes both. It is obvious that conducting multilingual multimodal abstractive summarization is more challenging due to the more complex scene~\cite{DBLP:journals/corr/abs-2109-05199}. Furthermore, our MM-Sum includes 44 languages, covering three settings: mid-high resource, low resource, and zero resource. What is more, our MM-Sum has the property that the knowledge can be transferred for MAS from mid-high-resource languages to low- and zero-resource languages via additional visual features as a bridge while they have not.

\begin{table}[t!]
\centering
\scalebox{0.52}{
\setlength{\tabcolsep}{0.150mm}{
\begin{tabular}{l|ccc}
\hline
\multirow{2}{*}{\textbf{Datasets}} &\multirow{2}{*}{\textbf{Num. of Lang.}} &\multirow{2}{*}{\textbf{Modalities}}&\multirow{2}{*}{\textbf{Scenes}}\\ 
 &&&\\
\hline
SportsSum~\cite{5711541} &1  &T,V,A   &Sports Video  \\
MovieSum~\cite{6527322} &1  &T,V,A   &Movies  \\
{MSMR}~\cite{1221239}   &1  &T,V   &Meeting Records  \\
{MMSS}~\cite{li-etal-2017-multi}  &2 &T,V,A &Multimedia   \\
{MSS}~\cite{li2018multi}  &1 &T,V  &Sentence  \\
{How2}~\cite{sanabria18how2}  &1   &T,V,A &YouTube Video  \\
{MSMO}~\cite{zhu-etal-2018-msmo}  &1   &T,V &News  \\
{E-DailyMail}~\cite{chen-zhuge-2018-abstractive}  &1 &T,V  &DailyMail Video \\
{EC-product}~\cite{li2020aspect}  &1   &T,V &E-Commerce Products \\
MM-AVS~\cite{fu-etal-2021-mm}  &1   &T,V,A &CNN\&DailyMail Video  \\
\cdashline{1-4}[4pt/2pt]
MultiLing2015~\cite{giannakopoulos-etal-2015-multiling}  &38 &T &Wikipedia\\
{GlobalVoices}~\cite{nguyen-daume-iii-2019-global}  &15 &T &News \\
{MultiSumm}~\cite{Cao_Wan_Yao_Yu_2020}  &2 &T  &News  \\
{MLSUM}~\cite{scialom-etal-2020-mlsum}  &5   &T &News \\
{MultiHumES}~\cite{yela-bello-etal-2021-multihumes}  &3   &T &Humanitarian Response \\
{MassiveSumm}~\cite{varab-schluter-2021-massivesumm}  &92 &T  &News \\
{MLGSum}~\cite{wang-etal-2021-contrastive}  &12   &T &News  \\
{XL-Sum}~\cite{hasan-etal-2021-xl}  &44   &T &News  \\
\cdashline{1-4}[4pt/2pt]
{MM-Sum (Ours)}  &44   &T,V &News  \\
\hline
\end{tabular}}}
\caption{Comparison of (1) previous multimodal abstractive summarization, (2) multilingual abstractive summarization, and (3) our MM-Sum. T/V/A: text/vision/audio modality. } 
\label{data:related}
\end{table}

\end{document}